# OBESITY HEURISTIC, NEW WAY ON ARTIFICIAL IMMUNE SYSTEMS

M. A. El-Dosuky[1], M. Z. Rashad[1], T. T. Hamza[1], and A.H. EL-Bassiouny[2]

[1] Department of Computer Sciences, Faculty of Computers and Information sciences, Mansoura University, Egypt
mouh_sal_010@mans.edu.eg
magdi_12003@yahoo.com
Taher_Hamza@yahoo.com

[2] Department of Mathematics, Faculty of Sciences, Mansoura University, Egypt
el_bassiouny@mans.edu.eg

## ABSTRACT

*There is a need for new metaphors from immunology to flourish the application areas of Artificial Immune Systems. A metaheuristic called Obesity Heuristic derived from advances in obesity treatment is proposed. The main forces of the algorithm are the generation omega-6 and omega-3 fatty acids. The algorithm works with Just-In-Time philosophy; by starting only when desired. A case study of data cleaning is provided. With experiments conducted on standard tables, results show that Obesity Heuristic outperforms other algorithms, with 100% recall. This is a great improvement over other algorithms.*

## 1. INTRODUCTION

Metaheuristics are global heuristic optimizers, usually inspired by nature [40], such as Artificial Immune Systems [11], genetic algorithms [16], ant colony optimization [10], particle swarm optimization [25] and Artificial Bee Colony [22]. All metaheuristics apply two strategies of ***intensification*** and ***diversification*** in effective exploration of search space [5]. Intensification eliminates the search space by examining neighbors of elite solutions, while diversification is a stochastic component that widens the search space by examining unvisited regions [15].

The diverse research into Artificial Immune Systems is surveyed by listing the application areas [17]. The major application areas are: Clustering/Classification, Anomaly Detection, Computer Security, Numeric Function Optimization, Combinatoric Optimization, and Learning. Minor application areas are: Bio-informatics, Image Processing, Control, Robotics, Virus Detection, and Web Mining.

Now, let us specify the problem statement as follows. There is an emphasis on the need for a support of a long-lasting extraction of new metaphors from immunology to flourish the application areas of Artificial Immune Systems [17].

We develop a new heuristic called Obesity Heuristic that is inspired from advances in medicine, especially in anti-Inflammatory treatment of obesity. **Section 2** reviews main Artificial Immune Systems notions and algorithm. **Section 3** introduces the underpinning principles for Obesity Heuristic. **Section 4** is the proposed algorithm. **Section 5** provides a case study of data cleaning in a typical data warehouse.

## 2. ARTIFICIAL IMMUNE SYSTEMS

Artificial Immune Systems algorithms are inspired by the studies of human Immune System [23]. Different types of proteins in human body can be classified into "self" and "non-self". T-cells are called "detector" cells, derived by the thymus, opposed to B-cells derived by bone marrow. It was noted that the immune system has a "memory" to remember illness-causing proteins (antigens) [23]. Ideas from human immune systems are applied in network security, especially intrusion detection system ([13], [35])

Clonal selection algorithm ([7], [8]) is described as follows:

1. Generate initial antibodies, where each antibody is a solution
2. Compute the fitness of each antibody.
3. Select highest fitness antibodies from population to generate new antibodies.
4. For each antibody, generate clones and mutate each clone.
5. Lower fitness antibodies are deleted from the population
6. New generated antibodies are added to the population
7. Repeat steps from 2- 6 until stop criterion is met. The stop criterion can be The number of iterations

## 2. OBESITY HEURISTIC

The direct relation among nutrition, infection and immunity seems to be intuitive. It took time to find the proof of the relation between inflammation and metabolic consequences of obesity, by clarifying the molecular mechanisms of inflammatory processes at that take place at the genetic level ([30], [37], [39]).

To facilitate describing the proposed heuristic, let us first review the obesity, processes involved in anti-Inflammatory treatment of obesity.

### 3.1 Obesity and the Perfect Nutritional Storm

The term "Perfect Nutritional Storm" [29] refers to the major changes in dietary. First, the increased consumption of refined carbohydrates notably increases the glycemic load of the diet, defined as the amount of consumed carbohydrate multiplied by the glycemic index [26]. Second, the increased consumption of refined vegetable oils rich in ***omega-6*** fatty acids is due to hormonal factors influencing desaturation by enzymes delta-6 and delta-5, strongly activated by insulin [6]. The third major change in dietary is the decreased consumption of long-chain ***omega-3*** fatty acids [34].

Obesity is the accumulation of excess body fat safely stored in ***adipose tissue*** in the form of triglycerides for long-term storage [24]. However, if the excess fat continues to increase and become deposited in any organ other than adipose tissue, this is known as ***lipotoxicity***, and this induces chronic diseases [36].

Insulin signaling is a main player in fat circulation, to provide the sufficient glucose levels into the fat cell that can be converted to glycerol [27].

### 3.2 Innate Immune System

Burns, injuries, microbial invasion are factors that can trigger inflammation through the innate immune response, now we know that and obese diet can do, as the innate immune system is highly evolved to become very sensitive to nutrients [21]. Advances in molecular biology shed light on inherent mechanisms of a typical innate immune system [20].

Inflammatory process is a function of two supposedly-balanced phases of pro- and anti-inflammatory phases. **Pro-inflammatory** mechanisms are indicators of cellular destruction, such as pain, swelling, redness and heat. **Anti-inflammatory** mechanisms are responsible for cellular repair and regeneration [33].

We may link pro- and anti-inflammatory mechanisms with intensification and diversification strategies mentioned in section 1.

Now, we need to show how our awareness of Perfect Nutritional Storm principle mentioned in section 3.1 can affect the behavior immune system.

A common *omega-6* fatty acid is Arachidonic acid (AA) and can induce inflammatory responses, so it must be reduced [19].

The explanation of the role of AA is as follows:

- Necrosis or cell death of a fat cell can occur, when the accumulation of AA in that fat cell increases over a threshold [31].

- This causes a migration of macrophages into the adipose tissue [38]

- These macrophages causes inflammatory mediators, such as IL-1, IL-6 and TNFα ([12] , [32])

*Omega-3* fatty acids, which is said to have an impact on brain development and intelligence [18], can activate anti-inflammatory gene transcription factors, so it must be increased enough [28].

## 4. PROPOSED OBESITY HEURISTIC

The proposed metaphor depends on obesity and immune system. The main forces of the algorithm are the generation *omega-6* and *omega-3* fatty acids. Within a large data set, we may think of raw data as simple amino acids and generated information and mined rules as complex fatty acids. Both omega-6 and omega-3 are generated and stored. Roughly speaking, *omega-6* can be a metaphor for undesired generated information and *omega-3* to be a metaphor for intelligent desired generated information. If the amount of *omega-6* is above a certain threshold, this is considered inflammation sign, and requires the start of immune system mechanisms to handle.

*Adipose tissue* storing triglycerides can be thought of as a metaphor for long-term data storage. If the excess fat continues to increase and become deposited in any location other than adipose tissue, this is known as *lipotoxicity*, and this induces malicious behaviors such as denial-of-service [4].

The hybrid method is described as follow:-

1. *Initialize Adipose tissues, the long-term data storage locations, to be empty*

2. Get the sequence of simple amino acids, the raw data from input streams

3. Generate complex fatty acids

4. *Store generated fatty acids*

5. If the storage location is not in one adipose tissue

    a. *If the size of this storage location is above threshold (**lipotoxicity**)*

        ➔ **start immune system mechanisms** to handle

6. Calculate the amount of generated **omega-6** and **omega-3** fatty acids

7. If the amount of **omega-6** is above a certain threshold,

    ➔ **start immune system mechanisms** to handle

8. *Repeat step 2- 7.*

9. *The performance measure of the system is* amount of generated **omega-3**

The main feature of the proposed algorithm is that it operates with Just-In-Time philosophy, by starting with full power only when desired as in $5^{th}$ and $7^{th}$ steps. We can utilize any AIS algorithm, but we prefer Clonal Selection Algorithm. In step 3, to generate complex fatty acids, we can apply any mining algorithm. However, we can assume that complex fatty acids are the amino acids, i.e., the output is the input for simplicity. This leads us for the following case study of data cleaning.

## 5. DATA CLEANING, A CASE STUDY

With Data Warehouse is an **integrated, subject-oriented, time-variant and non-volatile** database [46]. Integrating data from two or more heterogeneous sources shall cause some inevitable "*dirt*". In order to improve data consistency by reducing data redundancy, data cleaning is required [42]. Data cleaning is the automated detection and correction of missing and incorrect values [49].

Many methods are proposed for eliminating duplicate records in large data, such as sorting [41], sorting and clustering using a scanning window [45], basic field matching algorithm [48], pre-processing by tokenizing fields [47], and declarative language for data cleaning [43]. Simple performance measures can be:

$$\text{Recall} = \frac{|\text{correctly identified duplicates}|}{|\text{identified duplicates}|} \quad (1)$$

$$\text{False positive error} = \frac{|\text{wrongly identified duplicates}|}{|\text{identified duplicates}|} \quad (2)$$

Experiments are conducted using data warehouse with tables listed in [50].

Results show that Obesity Heuristic outperforms other algorithms, with a recall close to 100%, or false positive error close to zero. This is a great improvement over the algorithms described in [44].

## 6. CONCLUSIONS

We have introduced Obesity Heuristic a new metaphor derived from anti-Inflammatory treatment of obesity to the meet the need for a support of a long-lasting extraction of new metaphors from immunology to flourish the application areas Artificial Immune Systems. We believe that may extend the meaning and the application areas of Artificial Immune Systems. We review main Artificial Immune Systems notions and algorithm. Then we introduce the underpinning principles for Obesity Heuristic. We see that the proposed algorithm is promising especially in long-term huge data storage. Finally, the main feature of the proposed algorithm is that it links artificial immune system to huge storage such as data warehouses and makes it operates with Just-In-Time philosophy, by starting with full power only when desired as in 5th and 7th steps.

Future work may be the application of the proposed algorithm in other areas such as network security.

## ACKNOWLEDGEMENTS

The authors declare that they do not have any conflict of interest in the development of any phase of the submitted manuscript.

**Authors**

**Mohammed El-Dosuky** is a PhD student at Computer sciences Department, Faculty of Computers and Information, Mansoura University, Egypt. He received his Master in Computer Sciences about quantum computing from the Mansoura University in 2008. His current research interests are Multi-Agent Systems, intelligent communication, and evolutionary computing.

**Ahmed EL-Bassiouny** is currently a Professor at Mansoura University, Egypt. Currently, he is the dean of the Faculty of Sciences, Mansoura University.

**Taher Hamza** is currently an associate Professor at Mansoura University, Egypt. He is the vice dean For Higher Studies, Faculty of Computer and Information, Mansoura University.

**Magdy Rashad** is currently an associate Professor at Mansoura University, Egypt. He is the head of Computer Sciences department and the vice dean for Community Service and Environmental Development, Faculty of Computer and Information, Mansoura University.